\documentclass[letterpaper, 10 pt, conference]{ieeeconf}  

\IEEEoverridecommandlockouts                              

\usepackage{graphics} 
\usepackage{epsfig} 
\usepackage{mathptmx} 
\usepackage{times} 
\usepackage{amsmath} 
\usepackage{amssymb}  
\usepackage{algorithm} 
\usepackage{cite}
\usepackage{algpseudocode} 
\usepackage{lipsum}
\usepackage{graphicx}
\usepackage[dvipsnames]{xcolor}
\usepackage{eucal}
\usepackage{caption}
\usepackage{subcaption}
\usepackage{wrapfig}
\usepackage{cite}

\usepackage{multirow}

\usepackage{color, colortbl}
\usepackage[bookmarks=true]{hyperref}

\usepackage[utf8]{inputenc}
\usepackage{fontenc}
\usepackage{siunitx}
\usepackage{textcomp}
\usepackage{tikz}

\makeatletter
\newcommand*\titleheader[1]{\gdef\@titleheader{#1}}
\AtBeginDocument{%
  \let\st@red@title\@title
  \def\@title{%
    \bgroup\normalfont\large\centering\@titleheader\par\egroup
    \vskip1.5em\st@red@title}
}
\makeatother

\title{\LARGE \bf
Light-Weight Pointcloud Representation with Sparse Gaussian Process}
\author{Mahmoud Ali and Lantao Liu
\thanks{$^{1}$Mahmoud Ali and Lantao Liu are with the Luddy School of Informatics, Computing, and Engineering, Indiana University, Bloomington, IN 47408 USA, {\tt\small \{alimaa, lantao\}@iu.edu}}
}
\titleheader{ \textcolor{gray}{This paper has been accepted for publication at 2023 IEEE International Conference on Robotics and Automation} \textcolor{blue}{\href{https://www.icra2023.org/}{(ICRA 2023)}} }

\begin{document}

\maketitle
\thispagestyle{empty}
\pagestyle{empty}

\begin{abstract}
This paper presents a framework to represent high-fidelity pointcloud sensor observations for efficient communication and storage. 
The proposed approach exploits  Sparse Gaussian Process to encode pointcloud into a compact form. Our approach represents both the free space and the occupied space using only one model (one 2D Sparse Gaussian Process) instead of the existing two-model framework (two 3D Gaussian Mixture Models). We achieve this by proposing a variance-based sampling technique that effectively  discriminates between the free and occupied space. The new representation requires less memory footprint and can be transmitted across limited-bandwidth communication channels. The framework is extensively evaluated in simulation and it is also demonstrated using a real mobile robot equipped with a 3D LiDAR. Our method results in a 70$\sim$100 times reduction in the communication rate compared to sending the raw pointcloud. 

\end{abstract} 
\section{Introduction} 
With the rapid advancement of LiDAR technology, we now can build maps with remarkably high resolution. 
For example, each full scan of an only 16-channel  3D LiDAR  can give us 57600 points in the pointcloud that represents the surrounding obstacles. 
However, a price for using the high resolution LiDAR is the computation, storage, and communication costs when  mapping the environments.
While one might be able to upgrade the computation and storage by using a high performance computer system, the communication usually becomes a bottleneck due to the low communication bandwidth available.   
In practice,  the low bandwidth communication is considered as a major challenge for many robotics applications such as occupancy mapping of underwater and subterranean environments (caves, tunnels, mines, etc), search-and-rescue missions in disaster scenarios with a degraded communication infrastructure, and  planetary exploration missions~\cite{cid2018keeping}.  
The low bandwidth can prevent a robot from real-time sharing its sensor observations, and this can significantly degrade the system responsiveness if the robot needs to follow or interact with external control or supervision platforms.  
This work tackles the problem of sharing high-fidelity 3D pointcloud through a limited bandwidth communication channel. 

\begin{figure}[t] 
    \centering
    \input{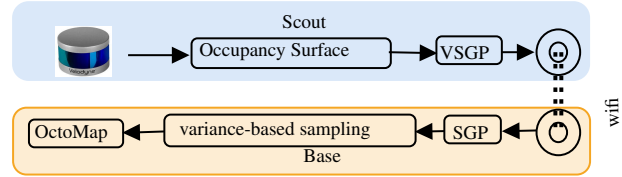}
    \caption{\small System Overview. \vspace{-5pt} }
    \label{block_diagram}    
\end{figure}
\begin{figure} [t]
    \centering
  \subfloat[\label{fig_entrance_a}]{%
      \includegraphics[width=0.33\linewidth,height=0.9in]{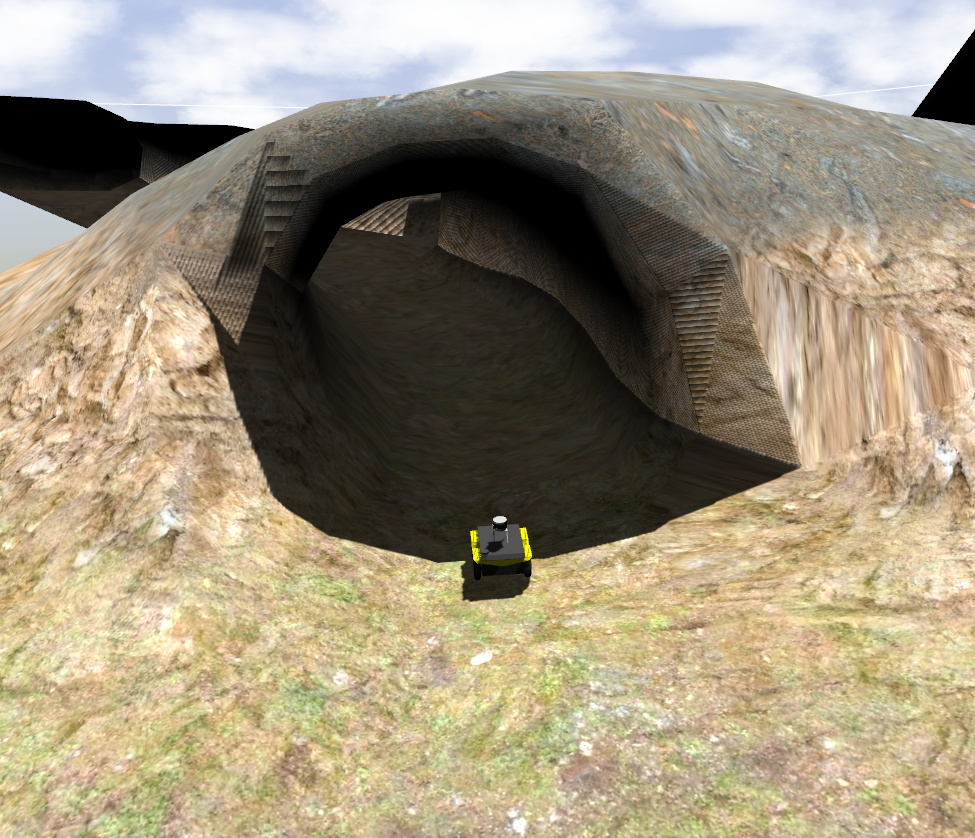}}
    \hfill
  \subfloat[\label{fig_entrance_b}]{%
      \includegraphics[width=0.33\linewidth,height=0.9in]{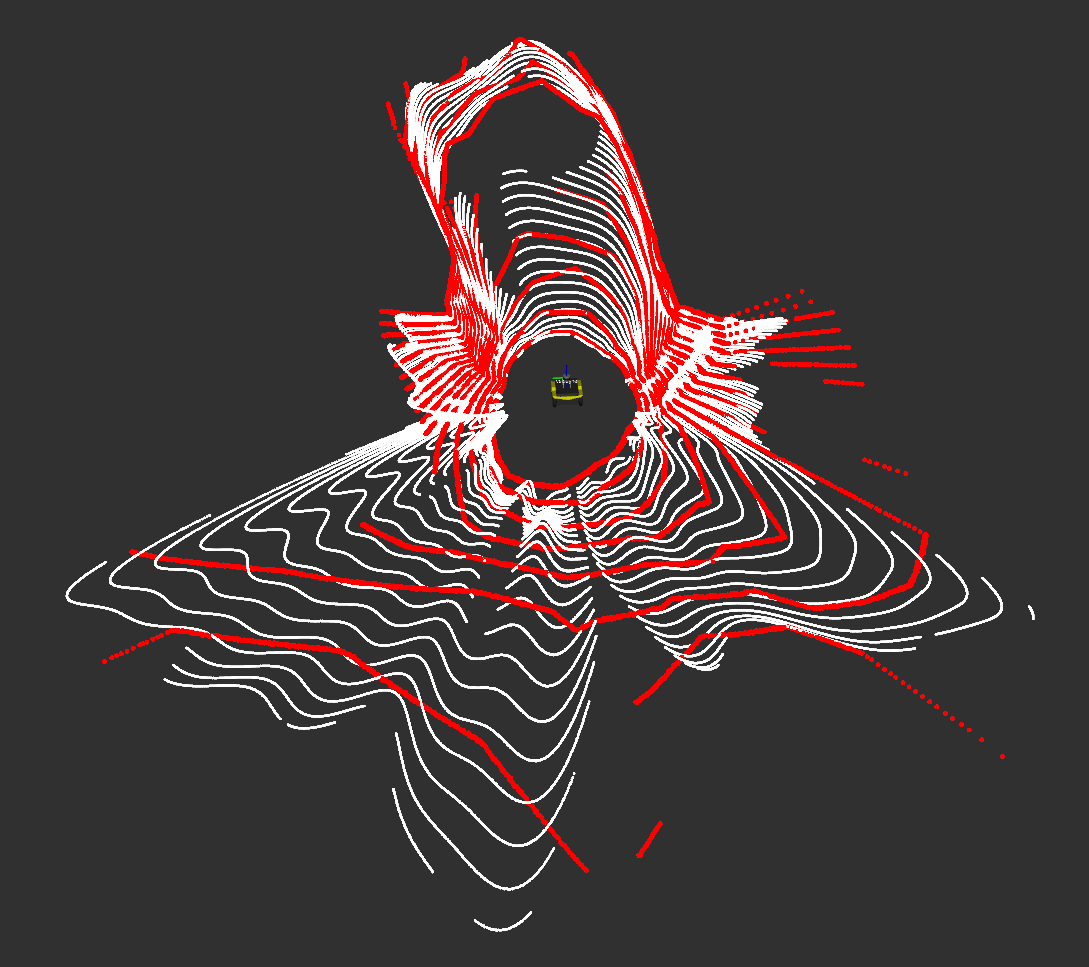}}
    \hfill
  \subfloat[\label{fig_entrance_c}]{%
      \includegraphics[width=0.33\linewidth,height=0.9in]{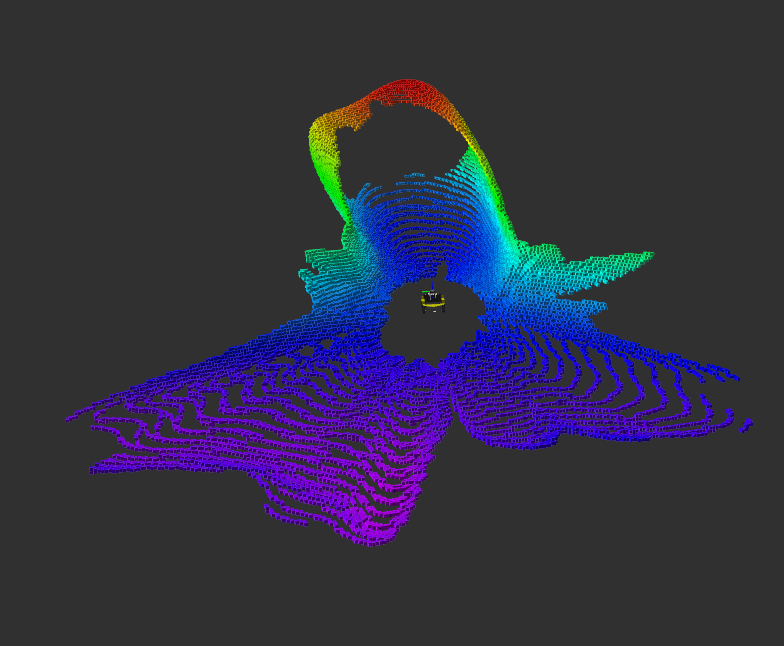}} \vspace{-5pt}
  \caption{\small 
  (a) Gazebo simulated mine tunnel; (b) Original pointcloud generated by a VLP16 LiDAR in red, and reconstructed pointcloud from the VSGP model in white; (d) Occupancy Map generated by OctoMap from the reconstructed pointcloud. \vspace{-10pt} 
  }
  \label{fig_entrance} 
\end{figure}

The system we consider consists of a robot (the scout) equipped with a LiDAR and a communication apparatus, and deployed in a low-bandwidth environment. The scout sends the observations that it acquires to a base for building the occupancy map of the environment, see Fig.~\ref{block_diagram}. 
Our approach exploits the {Variational Sparse Gaussian Process (VSGP)}~\cite{titsias2009variational} as a generative model to represent the pointcloud 
in a  compact form. 
This lightweight representation is transmitted through low-bandwidth communication to the base where the original pointcloud is reconstructed.
Extensive evaluations reveal that our approach results in a 70$\sim$100 times reduction in the memory as well as the communication rate required to transmit pointcloud data. 
%
For example, Fig.~\ref{fig_entrance_a} shows a scene of a simulated mine tunnel, where its raw pointcloud (shown in red, Fig.~\ref{fig_entrance_b}) requires around 750 KB of memory. Our approach is able to represent the same observation using only 6 KB of memory and transmit through limited-bandwidth communication. On the receiver side of the communication channel, the compact representation is used to reconstruct the original pointcloud (reconstructed pointcloud shown in white, Fig.~\ref{fig_entrance_b}). An occupancy map of the scene 
can be built using the reconstructed pointcloud, see Fig.~\ref{fig_entrance_c}.   

\section{Related Work} \label{related_work} 
Pointcloud compression algorithms have been investigated in recent years to cope with the demands to store and communicate the high-precision 3D points \cite{cao20193d}.
For example, the {space partitioning trees approaches} that exploit the 3D correlation between pointcloud points are widely used to compress the pointcloud data \cite{feng2020real, golla2015real, lasserre2019using, thanou2016graph, huang2006octree, huang2020octsqueeze}. 
Recently, {deep learning based approaches} were also proposed to leverage data and learn or encode the pointcloud compression \cite{quach2022survey, wiesmann2021deep,yan2019deep}. 
Different from these frameworks, the probabilistic approaches exploit the compactness of the distributions to compress 3D sensor observation. For instance, Gaussian Mixture Models (GMM)~\cite{realtime-expo, var-res-map-GMM, design2019communication} have been proposed as a generative model to encode 3D occupancy map. The GMM approach encodes the 3D data as a mixture of Gaussian densities to represent the occupied and free spaces around the robot.

Gaussian Process (GP) has been proven to be an excellent framework to model spatial phenomena or features in a continuous domain \cite{Rasmussen2005,Singh2007,Ouyang2014MAS}.  
Unfortunately, the standard GP has a cubic time complexity and this results in very limited scalability to large datasets. Methods for reducing the computing burdens of GPs have been previously investigated. For example, GP regressions can be done in a real-time fashion where the problem can be estimated locally with local data~\cite{Nguyen-tuong08localgaussian}. Sparse GPs (SGPs)~\cite{csato2002sparse,smola2001sparse,williams2001using,lawrence2003fast,snelson2006sparse,seeger2003bayesian,sheth2015sparse} tackle the computational complexity of the normal GP through leveraging the Bayesian rule with a sequential construction of the most relevant subset of the data.  


We propose a new probabilistic pointcloud compression approach which is based on the VSGP~\cite{titsias2009variational} and inspired by the GMM approach. While the GMM shares the accumulated sensory information as a set of accumulated Gaussian densities which are sampled and used as an occupancy map of the environment, in contrast, the proposed approach relies on sharing of immediate sensor observation to be reconstructed on the other side of the communication channel for further processing based on the required task (e.g. 3D mapping, object recognition, tracking, etc). 

This proposed VSGP-based approach offers a few advantages over the recent GMM approach:  while the {GMM} approach uses two {3D} GMMs to fit the occupied and free points~\cite{realtime-expo, var-res-map-GMM, design2019communication}, our approach uses only one {2D} {VSGP} to fit all the occupancy surface, including both the occupied and free points. 
The primary reason that our approach uses one VSGP instead of two is that we are using the variance calculated by the VSGP at each sampled point during the reconstruction process to decide if it belongs to the occupied or the free space. 
Therefore, {\em the proposed approach results in a more compact representation of the sensor observation, which requires less memory than the GMM approach and, as a consequence, leads to a lower communication rate.}


\section{Background} \label{background}
{GP} is a non-parametric model described by a mean function $m({x})$, and a co-variance function (kernel) $k({x}, {x}^{\prime})$, where ${x}$ is the {GP} input~\cite{GPforML}:\vspace{-1pt}
\begin{equation}
    f(\mathbf{x}) \sim \mathcal{G P}\left(m(\mathbf{x}), k\left(\mathbf{x}, \mathbf{x}^{\prime}\right)\right).
    \label{eq_full_gp}
\end{equation}
Considering a data set $\mathcal D = \left\{\left(\mathbf{x}_{i}, y_{i}\right)\right\}_{i=1}^{N}$ with $N$ training inputs $\mathbf{x}$ and their corresponding scalar outputs (observations) $\mathbf{y}$. 
After training the {GP} using the data set $\mathcal D$, the output $y_*$ for any new query ${x}_*$ can be estimated using the {GP} prediction:\vspace{-5pt}
\begin{equation}
    p(y_* | {y}) = N(y_* | m_{{y}}({x}_*), k_{y}({x}_*,{x}_*) + \sigma^2),
    \label{eq_predictive_eq_full_gp}
\end{equation} 
where $m_{\mathbf{y}}(\mathbf{x})$ and $k_{\mathbf{y}}({x}, {x}^{\prime})$ are the posterior mean and co-variance functions
~\cite{titsias2009variational}. The GP prediction equation depends on the values of the hyperparameters $(\Theta, \sigma^2)$ where $\Theta$ is the kernel parameters and $\sigma^2$ is the noise variance. 

The computation complexity of a full GP is $\mathcal{O}(N^3)$. 
In order to reduce the computation complexity, different approximation methods were proposed in the literature by considering only $M$ input points to represent the entire training data~\cite{GPforML}. These input points are called the {\em inducing points} $X_m$ and their corresponding values of the underlying function $f(\mathbf{x})$ are called the {\em inducing variables} $f_m$. Replacing the entire data set with only the M-inducing inputs leads to the SGP which has a computational complexity of $\mathcal{O}(NM^2)$. 
%
Titsias~\cite{titsias2009variational} proposed a variational learning framework to jointly estimate the kernel hyperparameters and the inducing points. Titsias' framework approximates the true exact posterior of a GP $p(f | y, \Theta)$ by a variational posterior distribution $ q(f,f_m)$, 
\begin{equation}
    q(f,f_m) = p(f|f_m)\phi(f_m),
    \label{eq_approx_equal_true_posterior_vsgp}
\end{equation}
where $\phi(f_m)$ is the free variational Gaussian 
distribution.
The Kullback-Leibler ($\mathbb{K} \mathbb{L}$) divergence is used to describe the discrepancy between the approximated and the true posteriors. Minimizing the  $\mathbb{K} \mathbb{L}$ divergence between the approximated and the true posteriors $\mathbb{K} \mathbb{L}[q(f, f_m)||p(f|y,\Theta)]$ is equivalent to maximizing the variational lower bound of the true log marginal likelihood: 
\vspace{-6pt}\begin{equation}
    \begin{gathered}
     F_{V}\left(X_{m}\right)=\log \left[N\left(\mathbf{y} \mid \mathbf{0}, \sigma^{2} I+Q_{n n}\right)\right]-\frac{1}{2 \sigma^{2}} \operatorname{Tr}(\widetilde{K}), \\
    Q_{n n}=K_{n m} K_{m m}^{-1} K_{m n}, \\
    \widetilde{K}=\operatorname{Cov}\left(\mathbf{f} \mid \mathbf{f}_{m}\right)= K_{n n}-K_{n m} K_{m m}^{-1} K_{m n},
    \end{gathered}
    \label{eq_Fv_Xm_vsgp}    
\end{equation}
 where $F_{V}\left(X_{m}\right)$ is the variational objective function,
$\operatorname{Tr}(\widetilde{K})$ is a regularization trace term,
$K_{nn}$ is the original $n \times n$ co-variance matrix,
$K_{mm}$ is $m \times m$ co-variance matrix on the inducing inputs,
$K_{nm}$ is $n \times m$ cross-covariance matrix between training and inducing points, and $K_{nm} = K_{mn}^T$. More details on VSGP can be found in Titsias's work \cite{titsias2009variational}. 
\vspace{-1pt}
\section{Methodology} \label{methodology} \vspace{-1pt}
The proposed approach exploits the VSGP as a generative model to encode 3D pointcloud. The VSGP is selected among different approximation approaches of GP due to the following reasons: 
i) The variational approximation distinguishes between the inducing points $M$ (as a variational parameter) and the kernel hyperparameters $(\Theta, \sigma)$. 
ii) The regularization term $\operatorname{Tr}(\widetilde{K})$ in the variational objective function (Eq.~\eqref{eq_Fv_Xm_vsgp}) regularizes the hyperparameters to avoid over-fitting of the data. 
iii) The variational approximation offers a discrete optimization scheme for selecting the inducing inputs $X_m$ from the original data\footnote{For more information about the inducing point selection, check \cite{titsias2009variational}}.
\subsection{{VSGP} as a generative model for the occupancy surface}
\begin{figure} \vspace{-5pt}
    \centering
  \subfloat[\label{fig_occ_surface_a}]{%
      \includegraphics[width=0.33\linewidth,height=1.0in]{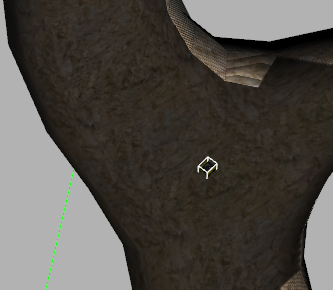}}
    \hfill
  \subfloat[\label{fig_occ_surface_b}]{%
      \includegraphics[width=0.33\linewidth,height=1.1in]{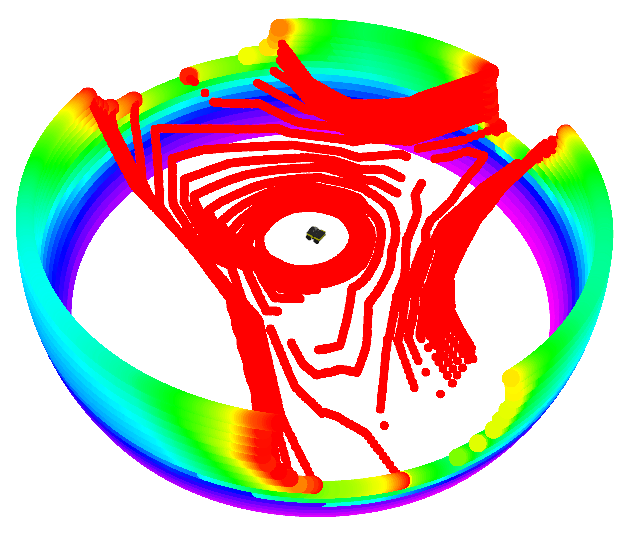}}
  \hfill
  \subfloat[\label{fig_occ_surface_c}]{%
      \includegraphics[width=0.33\linewidth,height=1.1in]{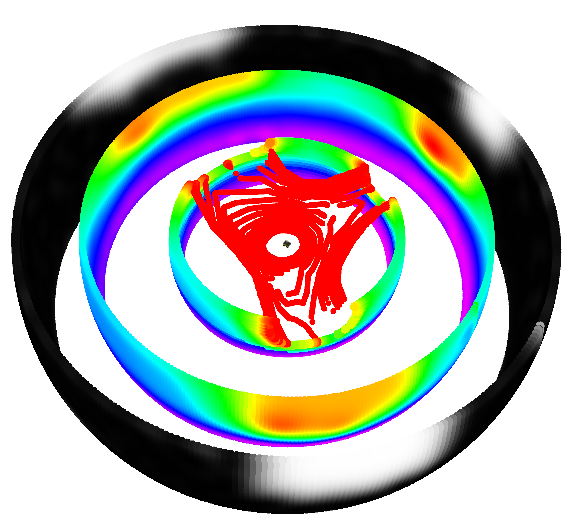}}
      \vspace{-5pt}
  \caption{\small 
  (a) Gazebo scene of a robot in a tunnel (black); (b) The occupancy surface generated from the original pointcloud, where warmer colors reflect smaller $f(\mathbf{x}_i)$ 
  values (less occupancy); (c) The inner surface represents the original occupancy surface (same as in b), and the middle surface represents the reconstructed occupancy surface using the VSGP model. The outer grey-coded surface represents the variance associated with each point on the reconstructed occupancy surface where brighter colors reflect high uncertainty. Raw pointcloud is shown in red in (b) and (c).\vspace{-10pt} }
  \label{fig_occ_surface} 
\end{figure}

Inspired by~\cite{realtime-expo}, we project the occupied points observed by a ranging sensor, e.g., LiDAR,  onto a circular surface around the sensor origin with a predefined radius $r_{oc}$. This surface is called {\em occupancy surface}, see Fig.~\ref{fig_occ_surface}. 
In our approach, the sensor observation is defined in the spherical coordinate system, 
where any {\em observed point} $x_i$ is described by the tuple $(\theta_i, \alpha_i, r_i)$ which represents the azimuth, elevation, and radius values, respectively. Also, any pointcloud data can be converted from the cartesian coordinates $(x_i, y_i, z_i)$ to the spherical coordinates $(\theta_i, \alpha_i, r_i)$ using the following equations:
\begin{equation}
    \begin{gathered}
    r_i = \sqrt{x_i^2 + y^2_i + z^2_i}, \ \ 
    \theta_i = \tan^{-1}(y_i, x_i), \ \ 
    \alpha_i = \cos^{-1}(z_i / r_i). 
    \end{gathered}
    \label{eq_cartesian_to_spherical}    
\end{equation}
All observed points that lie outside the circular occupancy surface (with a radius $r_i>r_{oc})$ or on the surface (with a radius $r_i=r_{oc})$ are neglected and considered as free space. The rest of the points that are inside the circular surface (with a radius $r_i<r_{oc})$ are projected on the occupancy surface and called the {\em occupied points}. Therefore, the occupancy surface radius $r_{oc}$ acts as the maximum range of the sensor. 
Each occupied point  ${x}_i$ on the surface is defined by two attributes: the azimuth and elevation angles ${x}_i= (\theta_i, \alpha_i)$, and assigned an {\em occupancy value} $f(\mathbf{x}_i)$ that is a function of the point radius $r_i$.  
The probability of occupancy $f(\mathbf{x}_i)$ at  each point on the occupancy surface is
modeled by a VSGP:
\vspace{-3pt}
\begin{equation}
    \begin{gathered}
    f(\mathbf{x}) \sim \mathcal{VSGP}\left(m(\mathbf{x}), k\left(\mathbf{x}, \mathbf{x}^{\prime}\right)\right).  \\
    \end{gathered}
    \label{eq_mean_kernel_vsgp}    
\end{equation}
 Considering noisy measurements, we add a white noise $\epsilon$ to the occupancy function $f(\mathbf{x})$, so the observed occupancy is described as  $y_i=f(\mathbf{x}_i)+\epsilon$
 where $\epsilon$ follows a Gaussian distribution ${N}\left(0, \sigma_{n}^{2}\right)$. 
 The final model of the occupancy surface is a 2D VSGP where the input is the azimuth and elevation angles, $ {x} \in  \{(\theta, \alpha)\}_{i=1}^n$, and the corresponding output is the expected occupancy $y_i$. The three main components of the final VSGP are:
\subsubsection{Zero-Mean Function $m({x})$}
There are different formulas to describe the relationship between the occupancy of a point $f(\mathbf{x}_i)$ on the occupancy surface and its radius $r_i$~\cite{realtime-expo}. For example, one candidate is $f(\mathbf{x}_i)=1/r_i$ where $r_i$ is bounded by the minimum and the maximum range of the sensor $r_{min}<r_i<r_{max}=r_{oc}$, where $r_{min}>0$.  
Our approach relates the occupancy of a point  $f(\mathbf{x}_i)$ to its radius $r_i$ by the following equation $f(\mathbf{x}_i)=r_{oc}-r_i$. This mapping between the occupancy and the radius of a point is compatible with the previous assumption that the occupancy surface radius $r_{oc}$ represents the maximum range of the sensor. Moreover, this mapping is encoded in our VSGP model as a zero-mean function $m(\mathbf{x})=0$ that sets the occupancy value of the non-observed points to zero. This mapping behavior mimics the mechanism of the LiDAR itself. 
\subsubsection{Rational Quadratic (RQ) Kernel}
The RQ kernel is selected because a GP prior with an RQ kernel is expected to have functions that vary across different length scales. This quality of the RQ kernel copes with the nature of the occupancy surface, specifically in unstructured environments where a range of diverse length scales is required, i.e., 
 \vspace{-5pt}
\begin{equation}
    \begin{gathered}
k_{\mathrm{RQ}}\left(\mathbf{x}, \mathbf{x}^{\prime}\right)=\sigma^{2}\left(1+\frac{\left(\mathbf{x}-\mathbf{x}^{\prime}\right)^{2}}{2 \alpha \ell^{2}}\right)^{-\alpha},
    \end{gathered}
    \label{eq_rq_krnl}    
\end{equation}
where $\sigma_{f}^{2}$ is the signal variance, $l$ is  the length-scale, and $\alpha$ sets the relative weighting of large and small scale variations.
The {RQ} co-variance function is more expressive in terms of modeling the occupancy surface than the most commonly used Squared Exponential (SE) co-variance function. This can be reasoned by the fact that the {RQ} kernel (when $\alpha$ and $l$ $>0$) is equivalent to a scale mixture of SE kernels with mixed characteristic length-scales~\cite{GPforML}. In practice, we take into account the resolution of LiDAR along both the azimuth and elevation axes to initiate different length-scales along each axis to reflect the LiDAR resolution. 
\subsubsection{Inducing Points Selection}
The variational learning framework proposed in~\cite{titsias2009variational} jointly optimizes the  variational parameters (inducing points) and the hyperparameters ($\Theta, \sigma$) through a variational Expectation-Maximization (EM) algorithm. In general, the original discrete optimization framework~\cite{titsias2009variational} suggests having an incremental set of the inducing points, so that during the Expectation step (E-step) a point from the input data is added to the inducing points set to maximize the variational objective function $F_{V}$ and minimize the $\mathbb{K} \mathbb{L}$ divergence between the true and approximated posteriors $\mathbb{K} \mathbb{L}[q(f)||p(f|y,\Theta)]$. Then the hyperparameters are updated during the Maximization step (M-step). 

Since LiDAR's field of view is limited within a certain range, the projection of the observed points on the circular surface leads to a limited input domain for the VSGP. In our case,  the azimuth and the elevation axes are limited to ($-\pi$ to $\pi$) and ($-15^{\circ}$ to $15^{\circ}$), respectively. The limited input domain is used to initiate a {\em fixed number of inducing points} at evenly distributed locations on the {\em occupied part} of the occupancy surface. 
In this way, a different combination of the points is selected at each E-step to maximize the variational objective function $F_{V}$ and minimize the  $\mathbb{KL}$ divergence. Then the hyperparameters are updated during the M-step. 
The number of the inducing points M is chosen to compromise the computational and memory complexity on one side and the accuracy of the reconstructed pointcloud on the other side. 
More inducing points result in higher computations complexity $\mathcal{O}(NM^2)$, larger memory to store the encoded observation, and higher bandwidth to transfer it. 
However, more inducing points increase the accuracy of the reconstructed pointcloud. 
We chose M=500 inducing points to keep the average deviation between the reconstructed pointcloud and the original pointcloud under 15 cm, see Section~\ref{sim_results} and Fig.~\ref{fig_metrics}. 
After the training phase on the scout side is completed, the selected inducing points are combined together with the hyperparameters values of the VSGP and are transmitted from the scout to the base. 
 \vspace{-5pt}
\subsection{Variance-based sampling} \label{sec_variance-based_sampling}
On the base side, the inducing points and the values of the hyperparameters, which are received from the scout, are used to reconstruct the original occupancy surface. The reconstruction is done through a {GP} configured with the same kernel (RQ) and likelihood (Gaussian) as the VSGP on the scout side. 
The base GP is trained on the inducing points and has a computation complexity of $\mathcal{O}(M^3)$ where $M$ is the number of the inducing points, so we refer it as a sparse GP (SGP) and refer the reconstructed occupancy surface as the {\em SGP occupancy surface}. A grid of query points ${x}_* = \{(\theta, \alpha)\}_{i=1}^K$ with the same resolution of the LiDAR along the azimuth and the elevation axes is generated to reconstruct the original pointcloud from the {SGP} occupancy surface -- we refer the reconstructed pointcloud as the {\em SGP pointcloud}. 
If up-sampling of the pointcloud is required for any reason, a query grid with higher resolution can be used for the reconstruction process.
The SGP occupancy surface is used to predict the occupancy $f({x}_i)$ of each point ${x}_i$ of the query grid ${x}_*$. The occupancy is converted back to the spherical radius  $r_i=r_{oc}-f(\mathbf{x}_i)$ to restore the 3D spherical coordinates of each point. 

\begin{figure}  \vspace{-5pt}
    \centering
  \subfloat[\label{fig_sampling_a}]{%
      \includegraphics[width=0.33\linewidth,height=1.1in]{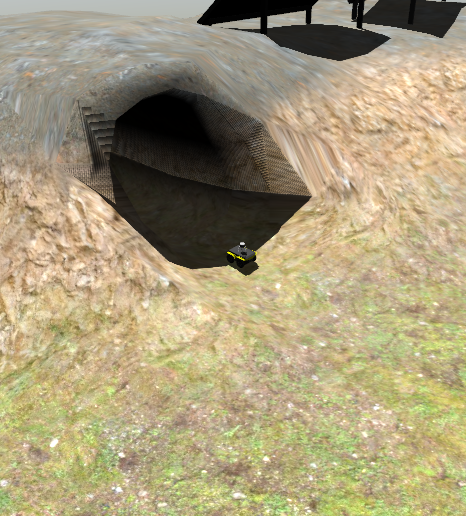}}
    \hfill
  \subfloat[\label{fig_sampling_b}]{%
      \includegraphics[width=0.33\linewidth,height=1.1in]{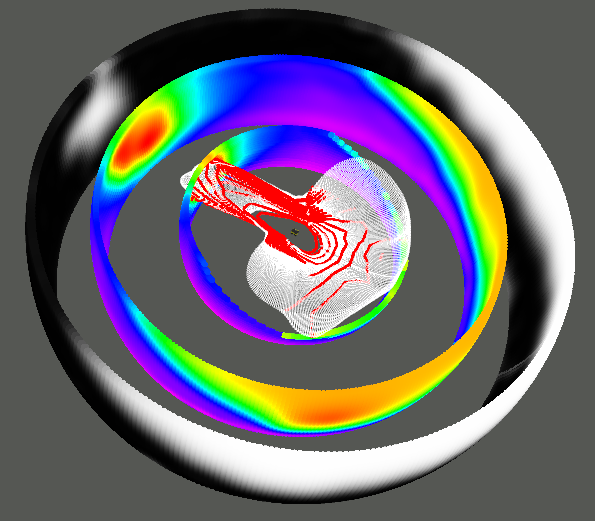}}
    \hfill
  \subfloat[\label{fig_sampling_c}]{%
      \includegraphics[width=0.33\linewidth,height=1.1in]{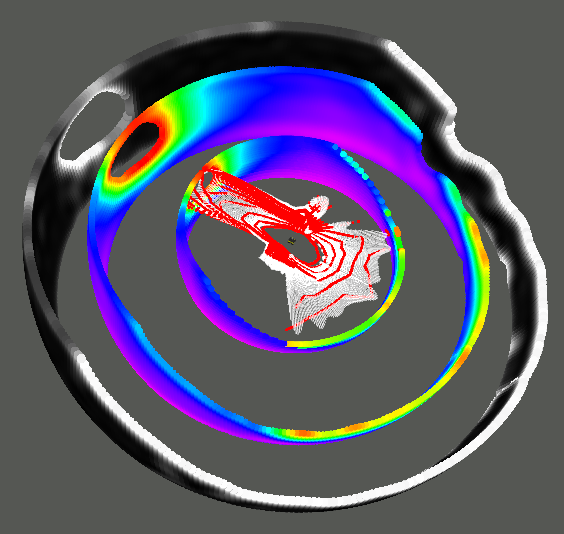}}
      \vspace{-5pt}
  \caption{ \small Variance-based sampling. (a) Gazebo scene shows the entrance of the tunnel; (b) shows the original (inner), reconstructed (middle), and variance (outer) surfaces. It also shows the reconstructed pointcloud (in white) through reconstructing from all points (free and occupied) of the occupancy surface. (c) shows reconstructed SGP pointcloud after removing all points that most likely belong to the free part of the occupancy surface. Raw pointcloud is shown in red in (b) and (c).
  \vspace{-10pt}}
  \label{fig_sampling} 
\end{figure}

One advantage of the GP and its variants over other modeling techniques is the uncertainty (variance) associated with the predicted value at any query point. Considering the VSGP model of the occupancy surface on the scout side, the variance associated with the occupied points is low compared to the variance related to the free points. Selecting the inducing points as a set from the original occupied points maintains low-variance values for the occupied part of the reconstructed SGP occupancy surface on the base side. Therefore, the variance value associated with any point on the reconstructed SGP occupancy surface is used to predict if that point belongs to the {\em occupied} or the {\em free} part of the occupancy surface, see Fig.~\ref{fig_sampling}.
We use a variance threshold $V_{th}$ as a judging criterion.  
In fact, the variance related to the occupancy surface is different from one observation to another, and it is affected by both the number of observed (occupied) points and their distribution over the occupancy surface. Therefore, we chose the variance threshold $V_{th}$ as a variable that changes with the distribution of the variance over the occupied and free parts of the occupancy surface. $V_{th}$ is defined as a linear combination of the variance mean $v_m$ and standard deviation $v_{std}$ over the surface, i.e.,  $V_{th}= K_m*v_m + K_{std}*v_{std}$ where $K_m$ and $K_{std}$ are constants.
These two constants are tuned by first setting $V_{th} = v_m$ ($K_m=1$ , $K_{std}=0$), then we increase $K_{std}$ and decrease $K_m$ gradually till we get the values that give the highest accuracy for the reconstructed SGP pointcloud (considering a fixed number of inducing points). 
Our sampling-based approach is capable of discriminating between the free points that most likely belong to {the free part} of the SGP occupancy surface and the occupied points that belong to the {the occupied part} of the SGP occupancy surface. 
After removing {\em the free part} of the SGP occupancy surface, 
the Cartesian coordinates of the occupied points are calculated using the inverse form of Eq.~\eqref{eq_cartesian_to_spherical} to restore the original point cloud, see Fig.~\ref{fig_sampling_c}.

\section{Experimental Design and Results}
The proposed approach is implemented in Python3 on top of GPflow-v2~\cite{matthews2017gpflow} and TensorFlow-v2.4~\cite{abadi2016tensorflow} under ROS framework~\cite{quigley2009ros}. Both real-time simulation and real-time demonstration were considered to evaluate the proposed approach. In both the simulation and the hardware experiments, a VLP-16 LiDAR 
was used with a maximum range of $10 m$, a frequency of $4 Hz$, and a resolution of $(0.1^{\circ}, 2^\circ)$ along the azimuth and the elevation axis, respectively. This configuration results in a maximum pointcloud size of 57600 points. 
The query grid, which is used to sample the SGP occupancy surface on the base side, has the same resolution as the VLP-16 LiDAR. A 3D occupancy grid map with a resolution of $5 cm$ is generated from the reconstructed SGP pointcloud through Octomap~\cite{hornung2013octomap}. 

We investigate the performance of our framework and compare it with the GMM approach~\cite{realtime-expo, var-res-map-GMM, design2019communication}.
While the GMM approach tackles the occupancy mapping problem as a whole, 
our approach focuses on compressing sensor observations through limited-bandwidth communication channels. To be able to compare the two approaches, we implemented the GMM approach in such a way that it is used to encode one sensor observation at a time instead of generating an entire occupancy map. We compared our approach with two versions of the GMM approach: i) A CPU-based implementation of GMM that follows the same guidelines of~\cite{realtime-expo}. ii) An upgraded GPU-based implementation of GMM. We implemented the GPU-GMM to have a fair computation comparison with our VSGP approach which runs on GPU. 
\subsection{Simulation Experiments} \label{simulation_experiment}
\subsubsection{Simulation Setup:} \label{sim_setup}
The simulation setup consists of two machines that communicate to each other over WiFi: The first machine, where the scout and the environment are simulated, is an Intel® Core™ i7 {\em{ NUC11}} PC equipped with 64 GB RAM and 6 GB  Geforce $RTX2060$ GPU. The second machine, which acts as the base, is an Intel® Core™ i7 {\em {Alienware}} Laptop equipped with 32 GB RAM and 8 GB Geforce $RTX2080$ GPU. Both are connected using a 2.4 GHz WiFi router. The network flow is monitored using the {\em ifstat} tool to evaluate the communication performance. 
The mine tunnel of the {\em cpr\_inspection} world, which is developed by \textit{ClearPath} robotics, is used as our simulation environment. This environment is selected because it represents one of the targeted low-bandwidth subterranean environments. The mine tunnel part of the {\em cpr\_inspection} world fits in a rectangular area with an approximated area of $30\times65 m^2$, the tunnel length is around $135m$. The ground elevation and the height of the tunnel are different from one place to another. The ClearPath Jackal robot is used as the scout. 
The proposed approach was evaluated through 20 real-time simulation trials. In each trial, the robot starts at the beginning of the cave and follows a predefined path along the mine using way-point based navigation. 
\vspace{2pt}
\subsubsection{Simulation Results} \label{sim_results}
We evaluate the performance of our approach based on the reduction in the memory and the communication rate 
required to transmit the sensor observations between the scout and the base. The VSGP representation requires only  $1514$ floating points (FP) to represent the entire pointcloud (3 FP for each inducing point (3x500) + 6 FP for robot pose + 6 FP for the hyperparameters). This value is less than the memory needed by the GMM approach which requires $\sim 2000$ FP (10 FP for each component (10x200) distributed as 6 FP for covariance + 3 FP for mean + 1 FP for weight)~\cite{realtime-expo}. We send the robot pose to the base because our approach encodes the observation relative to the robot body frame, while the GMM approach first transforms the observation from the robot body frame to a global frame using the robot current pose and then sends the encoded Gaussians densties with respect to the global frame.

\begin{figure}[t] \vspace{-1pt}
    \centering
     \subfloat[\label{fig_metrics_a}]{%
      \includegraphics[width=0.49\linewidth,height=1.4in]{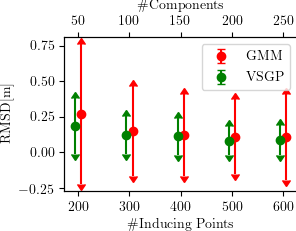}}
    \hfill
  \subfloat[\label{fig_metrics_b}]{%
      \includegraphics[width=0.49\linewidth,height=1.4in]{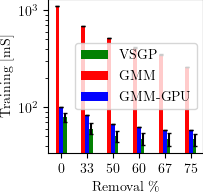}}
    \\
  \subfloat[\label{fig_metrics_c}]{%
      \includegraphics[width=0.47\linewidth,height=1.4in]{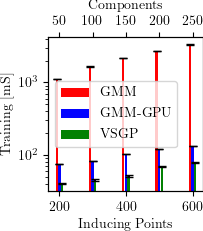}}
    \hfill
  \subfloat[\label{fig_metrics_d}]{%
      \includegraphics[width=0.49\linewidth,height=1.4in]
      {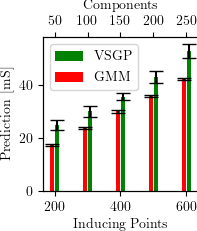}}
   \vspace{-5pt}
  \caption{\small Performance comparisons. (a) shows the RMSD between the reconstructed and the original pointcloud for VSGP(vs \#inducing points) and GMM(vs \#components);
  (b) illustrates the training time against the pointcloud size (considering 500-inducing points VSGP, and equivalently, 200-components GMM); (c) represents the training time versus the \#VSGP-inducing points and \#GMM-components; (d) shows the prediction time versus the \#VSGP-inducing points and \#GMM-components. \vspace{-15pt}}
  \label{fig_metrics} 
\end{figure}

To quantify the accuracy of the reconstructed SGP pointcloud, we use the Root Mean Square Deviation (RMSD) between the radius predicted by our approach and the actual radius of each point on the occupancy surface.
\vspace{-5pt}
\begin{equation}
    \begin{gathered}
\mathrm{RMSD}=\sqrt{\frac{\sum_{i=1}^{N}\left(r_{i}-\hat{r}_{i}\right)^{2}}{N}},
    \end{gathered}
    \label{eq_rmse}    
\end{equation}
where $N$ is the size of the pointcloud, 
$r_{i}$ is the actual radius at $(\theta_i, \alpha_i)$, and  
$\hat{r}_{i}$ is the estimated radius value at the same point $(\theta_i, \alpha_i)$. 
Fig.~\ref{fig_metrics_a} shows the mean and the standard deviation of the RMSD for each predicted point over $110$ observations (each observation has around $10K$ to $50K$ points). 
Also, Fig.~\ref{fig_metrics_a} implicitly reflects the memory required by VSGP and GMM to store one observation, as described before that the memory required to store one observation can be calculated by multiplying the number of inducing points (bottom x-axis) by 3 and multiplying the number of components (top x-axis) by 10. 
We match pairs of the VSGP and GMM models (in terms of the number of inducing points and components) based on the memory requirement and the accuracy of the reconstructed pointcloud (reflected by the RMSD) for each pair, see table \ref{table_1}. For example, $500$-inducing points VSGP results in an average RMSD value for each point of $9$ cm with a standard deviation of $10$ cm. This corresponds to an average RMSD of $11$ cm with a standard deviation of $25$ cm for a $200$-components GMM.


{\small
\captionof{table}{VSGP vs GMM(ind: inducing, cps: components)}
\begin{center}
\begin{tabular}{ c|c|c||c|c|c } 
 \multicolumn{3}{c||}{VSGP} & \multicolumn{3}{c}{GMM} \\
  \hline
\#  & Memory & RMSD  & \#  & Memory & RMSD\\
  ind  & $\sim$FPs    & $\sim$cm    & cps  & $\sim$FPs    & $\sim$cm\\
 \hline
 200 & 600 & 20$\pm$22     & 50 & 500    & 27$\pm$50 \\ 
 300 & 900 & 14$\pm$15     & 100 & 1000 & 16$\pm$35 \\ 
 400 & 1200 & 12$\pm$14    & 150 & 1500 & 13$\pm$31 \\ 
 500 & 1500 & 9$\pm$10     & 200 & 2000 & 11$\pm$29 \\ 
 600 & 1800 & 9$\pm$10   & 250 & 2500 & 11$\pm$30 \\ 
 \hline
\end{tabular}
\label{table_1}
\end{center}
}
\vspace{4pt}
Now we analyze the results in Fig.~\ref{fig_metrics}.  Fig.~\ref{fig_metrics_a} shows the RMSD values associated with VSGP have a smaller standard deviation than the 
GMM's. 
It also shows that increasing the number of the VSGP-inducing points (bottom x-axis) or the number of the GMM-components (top x-axis) will result in smaller RMSD (higher accuracy).

\begin{figure}[t]
    \centering
  \subfloat[\label{fig_final_map_a}]{%
      \includegraphics[width=0.33\linewidth,height=01.1in]{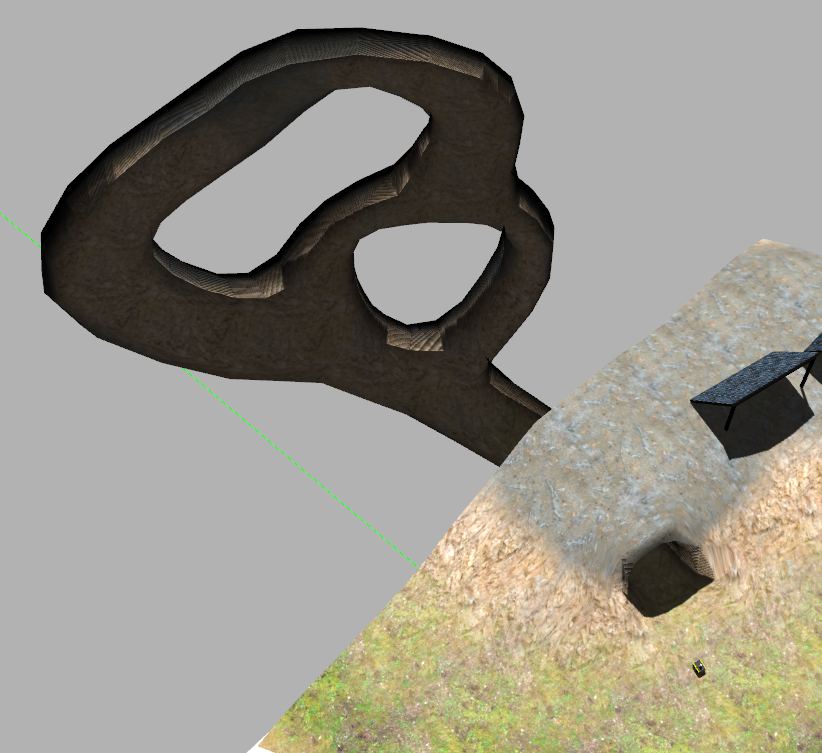}}
    \hfill
  \subfloat[\label{fig_final_map_b}]{%
      \includegraphics[width=0.33\linewidth,height=01.1in]{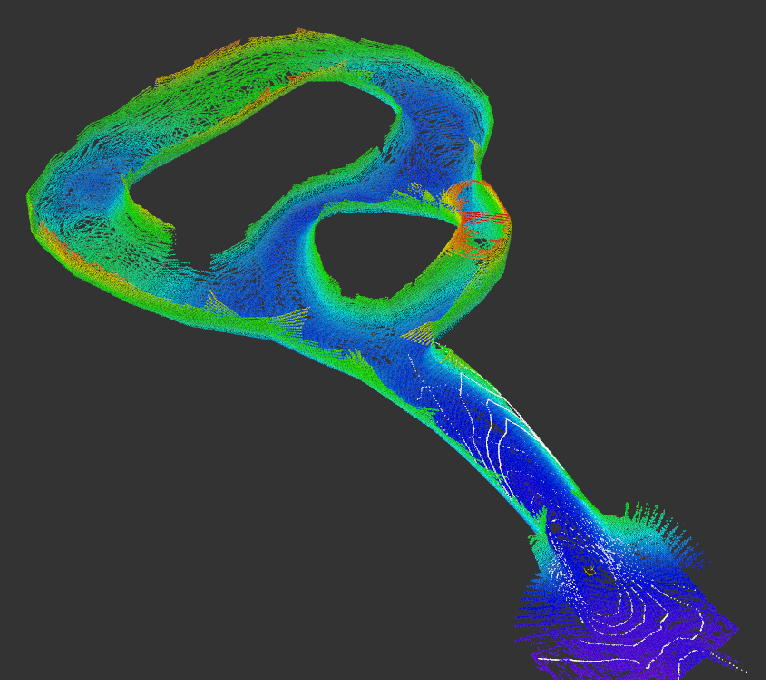}}
    \hfill
  \subfloat[\label{fig_final_map_c}]{%
      \includegraphics[width=0.33\linewidth,height=01.1in]{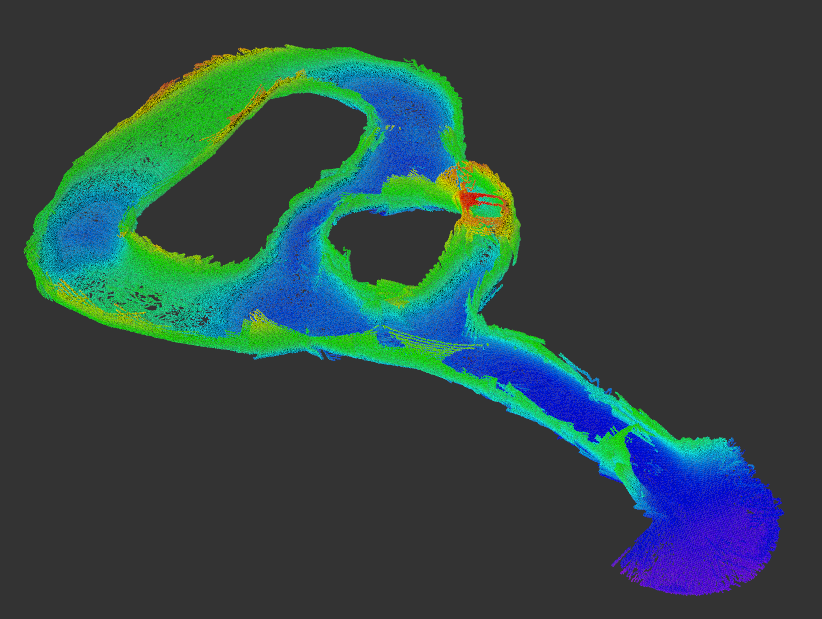}}
      \vfill
          \vspace{5pt} 
 \subfloat[\label{fig_final_map_d}]{%
      \includegraphics[width=\linewidth,height=1.6in]{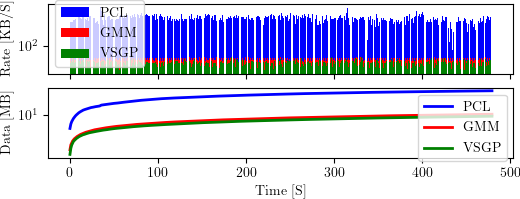}}\vspace{-5pt}
  \caption{\small 
  (a) shows the simulated mine environment in Gazebo; (b) shows the Octomap of the mine generated from the original pointcloud; (c) shows the Octomap generated from the reconstructed SGP pointcloud; (d) shows the communication rate  and the accumulated data sent from the scout to the base in case of sending raw pointcloud PCL(1750KB/S, 840MB), GMM data(25.8KB/S, 12.4MB), and VSGP data(18.2KB/S, 8.7MB). The y-axis is plotted in log-scale. \vspace{-10pt} } 
  \label{fig_final_map} 
\end{figure}

An intensive evaluation of the training and the prediction phases is presented in Figs.~\ref{fig_metrics_b}-\ref{fig_metrics_d}. The reduction in the training time versus the reduction in the size of the raw pointcloud is presented in Fig.~\ref{fig_metrics_b}, where $0\%$ removal percent means a pointcloud size of 57.6K points. Fig.~\ref{fig_metrics_c} shows the increase in training time versus the number of inducing points and the number of components. We compare the training time of the VSGP, the GMM-CPU (considering the default configuration of the GMM approach used in~\cite{realtime-expo}), and the GMM-GPU implementation. The results show that our approach outperforms both the CPU and GPU implementation of the GMM approach in terms of training time.
Fig.~\ref{fig_metrics_d} presents the variation of the prediction time of the VSGP versus the number of the inducing points, where the values shown in the figure represent the time required to predict the occupancy value associated with all the points of the grid query $\mathbf{x}$ (57600 points). 

Fig.~\ref{fig_metrics_d} indicates that for a matching pair of GMM and VSGP (Table \ref{table_1}), GMM has a less sampling time than the paired VSGP. However, the pointcloud reconstruction process of the VSGP is more convenient than the GMM approach because a fundamental difference between sampling the VSGP and the GMM is that: when we {sample} from a GMM, we get a sample (from a distribution) with random values $(\theta_s, \alpha_s, r_s)$, so we {\em do not have control} over the location of the sample on the occupancy surface $(\theta_s, \alpha_s)$. In contrast, for the VSGP approach, we {predict} the radius value $r_s$ for a certain point on the occupancy surface defined by $(\theta_s, \alpha_s)$. So, we {\em have control} over the point location on the occupancy surface. 
While constructing the 3D octomap of the tunnel environment using the scout-base scheme, the average communication rate was 1750 KB/S, 25.8 KB/S, and 18.2 KB/S for sending raw point clouds, GMM encoded data, and VSGP encoded data respectively, see Fig.~\ref{fig_final_map_d}. The accumulated data sent through the network is reduced from $840$ MB for sending raw pointcloud to 12.4 MB in case of GMM and $8.7$ MB in case of VSGP. This indicates a compression ratio of $\sim 96$ ($840/8.7 \sim 1750/18.2)$. 
\subsection{Hardware Experiment} \label{hardware_experiment} 
\begin{figure}[t]
    \centering
  \subfloat[\label{fig_mesh_map_a}]{%
      \includegraphics[width=0.495\linewidth,height=1.1in]{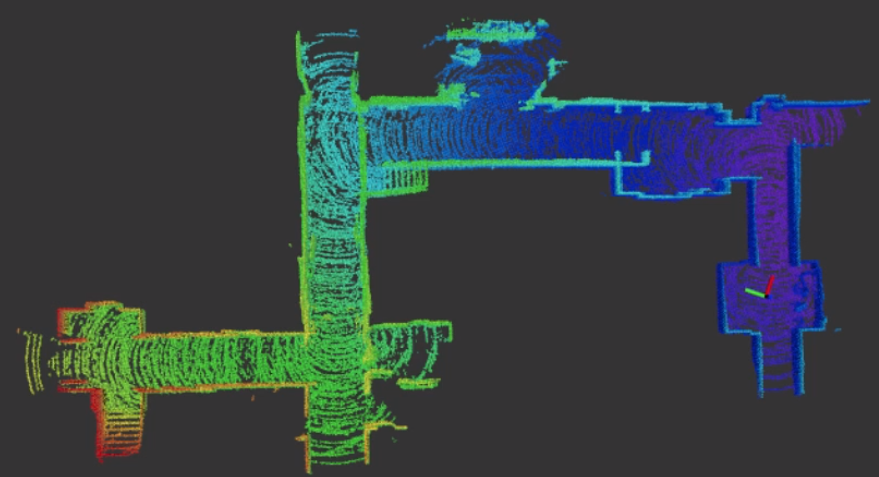}}
    \hfill
  \subfloat[\label{fig_mesh_map_b}]{%
      \includegraphics[width=0.495\linewidth,height=1.1in]{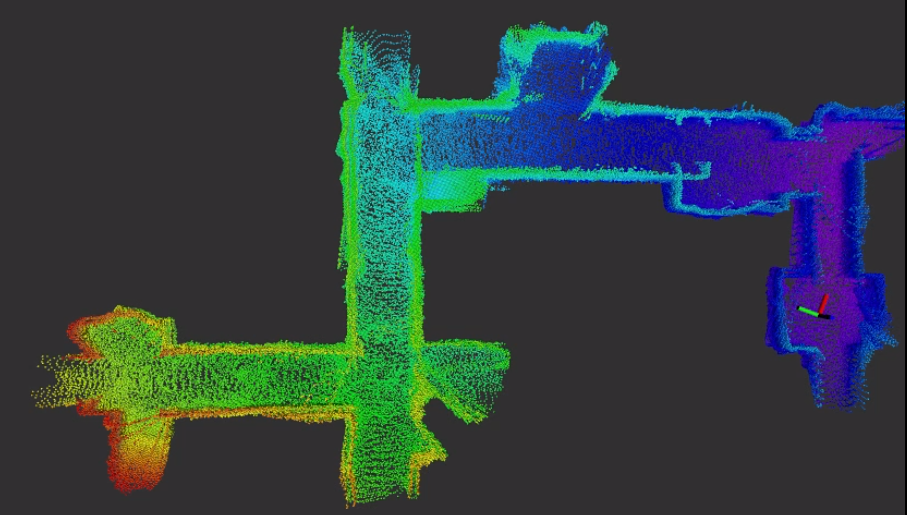}}
    \vfill
   \vspace{5pt} 
  \subfloat[\label{fig_mesh_map_c}]{%
      \includegraphics[width=\linewidth,height=1.6in]{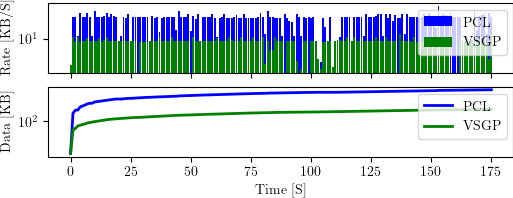} }
  \vspace{-5pt} 
  \caption{ \small Indoor demonstration. (a) shows octomap of the laboratory building generated from the original pointcloud. (b) shows octomap generated from the reconstructed SGP pointcloud. (c) shows the reduction in the communication rate and the accumulated data sent from the scout to the base, where log-scale is used for y-axis. PCL represents the raw pointcloud.\vspace{-10pt} }
  \label{fig_mesh_map} 
\end{figure}

A Jackal mobile robot, equipped with a VLP-16 LiDAR and {NUC11} PC, was used as the scout, while the {Alienware} laptop was used as the base. 
The demonstration was conducted in an indoor environment, where the VSGP-encoded pointcloud data was sent from the scout to the base to generate a 3D Octomap~\cite{hornung2013octomap} of the building from the SGP reconstructed pointcloud in real-time, see Fig.~\ref{fig_mesh_map}. 
Fig.~\ref{fig_mesh_map}c shows the reduction in the communication rate for the hardware experiment. The communication rate dropped from around 560 KB for transmitting raw pointcloud to around 8 KB for transmitting the encoded VSGP (this ratio is equivalent to $70$ times smaller rate). The communication rate of the hardware experiment is low compared to the simulation experiment because the LiDAR resolution was halved during the hardware experiment. The total amount of data transmitted at the end of each trial was around 100 MB for sending raw pointcloud and only around 1.4 MB for sending the VSGP encoded observation.  

\section{Conclusion} \label{conclusion}
In this paper, we introduce a lightweight representation for the 3D pointcloud using the VSGP. This representation allows high-fidelity observations to be efficiently stored and transmitted through limited-bandwidth communication channels. 
Based on the results of the simulation and hardware experiments, our approach results in around 70-100 times smaller size representation of the sensor observation. This compact representation can facilitate many of the robotics applications which are limited by the communication bandwidth such as subterranean and underwater exploration, search and rescue missions, and planetary exploration. In addition, our approach can also be beneficial in the context of multi-robot collaboration where a number of robots are required to share high-volume information (3D pointcloud) through low-bandwidth channels.

\section*{Acknowledgement}
This work was supported by National Science Foundation  with grant numbers 2006886 and 2047169.

\bibliographystyle{unsrt}
\bibliography{references,ref2,ref3}

\begin{thebibliography}{10}

\bibitem{cid2018keeping}
V.~H. Cid et~al.
\newblock Keeping communications flowing during large-scale disasters:
  leveraging amateur radio innovations for disaster medicine.
\newblock {\em Disaster medicine and public health preparedness},
  12(2):257--264, 2018.

\bibitem{titsias2009variational}
Michalis Titsias.
\newblock Variational learning of inducing variables in sparse gaussian
  processes.
\newblock In {\em Artificial intelligence and statistics}, pages 567--574.
  PMLR, 2009.

\bibitem{cao20193d}
Chao Cao, Marius Preda, and Titus Zaharia.
\newblock 3d point cloud compression: A survey.
\newblock In {\em The 24th International Conference on 3D Web Technology},
  pages 1--9, 2019.

\bibitem{feng2020real}
Yu~Feng, Shaoshan Liu, and Yuhao Zhu.
\newblock Real-time spatio-temporal lidar point cloud compression.
\newblock In {\em 2020 IEEE/RSJ international conference on intelligent robots
  and systems (IROS)}, pages 10766--10773. IEEE, 2020.

\bibitem{golla2015real}
Tim Golla and Reinhard Klein.
\newblock Real-time point cloud compression.
\newblock In {\em 2015 IEEE/RSJ International Conference on Intelligent Robots
  and Systems (IROS)}, pages 5087--5092. IEEE, 2015.

\bibitem{lasserre2019using}
S{\'e}bastien Lasserre, David Flynn, and Shouxing Qu.
\newblock Using neighbouring nodes for the compression of octrees representing
  the geometry of point clouds.
\newblock In {\em Proceedings of the 10th ACM Multimedia Systems Conference},
  pages 145--153, 2019.

\bibitem{thanou2016graph}
Dorina Thanou, Philip~A Chou, and Pascal Frossard.
\newblock Graph-based compression of dynamic 3d point cloud sequences.
\newblock {\em IEEE Transactions on Image Processing}, 25(4):1765--1778, 2016.

\bibitem{huang2006octree}
Yan Huang, Jingliang Peng, C-C~Jay Kuo, and M~Gopi.
\newblock Octree-based progressive geometry coding of point clouds.
\newblock In {\em PBG@ SIGGRAPH}, pages 103--110, 2006.

\bibitem{huang2020octsqueeze}
Lila Huang, Shenlong Wang, Kelvin Wong, Jerry Liu, and Raquel Urtasun.
\newblock Octsqueeze: Octree-structured entropy model for lidar compression.
\newblock In {\em Proceedings of the IEEE/CVF conference on computer vision and
  pattern recognition}, pages 1313--1323, 2020.

\bibitem{quach2022survey}
Maurice Quach, Jiahao Pang, Dong Tian, Giuseppe Valenzise, and Fr{\'e}d{\'e}ric
  Dufaux.
\newblock Survey on deep learning-based point cloud compression.
\newblock {\em Frontiers in Signal Processing}, 2022.

\bibitem{wiesmann2021deep}
Louis Wiesmann, Andres Milioto, Xieyuanli Chen, Cyrill Stachniss, and Jens
  Behley.
\newblock Deep compression for dense point cloud maps.
\newblock {\em IEEE Robotics and Automation Letters}, 6(2):2060--2067, 2021.

\bibitem{yan2019deep}
Wei Yan, Shan Liu, Thomas~H Li, Zhu Li, Ge~Li, et~al.
\newblock Deep autoencoder-based lossy geometry compression for point clouds.
\newblock {\em arXiv preprint arXiv:1905.03691}, 2019.

\bibitem{realtime-expo}
W.~Tabib et~al.
\newblock Real-time information-theoretic exploration with gaussian mixture
  model maps.
\newblock In {\em Robotics: Science and Systems}, 2019.

\bibitem{var-res-map-GMM}
C.~O’Meadhra et~al.
\newblock Variable resolution occupancy mapping using gaussian mixture models.
\newblock {\em IEEE Robotics and Automation Letters}, 4(2):2015--2022, 2018.

\bibitem{design2019communication}
Task-Specific~Manipulator Design.
\newblock Communication-efficient planning and mapping for multi-robot
  exploration in large environments.
\newblock {\em Journal Article}, 15(2):e1971, 2019.

\bibitem{Rasmussen2005}
Carl~Edward Rasmussen and Christopher K.~I. Williams.
\newblock {\em Gaussian Processes for Machine Learning}.
\newblock The MIT Press, 2005.

\bibitem{Singh2007}
A.~Singh, A.~Krause, C.~Guestrin, W.~Kaiser, and M.~Batalin.
\newblock Efficient planning of informative paths for multiple robots.
\newblock In {\em the 20th International Joint Conference on Artifical
  Intelligence}, IJCAI'07, pages 2204--2211, 2007.

\bibitem{Ouyang2014MAS}
Ruofei Ouyang, Kian~Hsiang Low, Jie Chen, and Patrick Jaillet.
\newblock Multi-robot active sensing of non-stationary gaussian process-based
  environmental phenomena.
\newblock In {\em Proceedings of the 2014 International Conference on
  Autonomous Agents and Multi-agent Systems}, pages 573--580, 2014.

\bibitem{Nguyen-tuong08localgaussian}
Duy Nguyen-tuong and Jan Peters.
\newblock Local gaussian process regression for real time online model learning
  and control.
\newblock In {\em In Advances in Neural Information Processing Systems 22
  (NIPS}, 2008.

\bibitem{csato2002sparse}
Lehel Csat{\'o} and Manfred Opper.
\newblock Sparse on-line gaussian processes.
\newblock {\em Neural computation}, 14(3):641--668, 2002.

\bibitem{smola2001sparse}
A.~J. Smola and P.~L. Bartlett.
\newblock Sparse greedy gaussian process regression.
\newblock In {\em Advances in neural information processing systems}, pages
  619--625, 2001.

\bibitem{williams2001using}
Ch. Williams and M.~Seeger.
\newblock Using the nystr{\"o}m method to speed up kernel machines.
\newblock In {\em Proceedings of the 14th annual conference on neural
  information processing systems}, number CONF, pages 682--688, 2001.

\bibitem{lawrence2003fast}
N.~Lawrence, M.~Seeger, and R.~Herbrich.
\newblock Fast sparse gaussian process methods: The informative vector machine.
\newblock In {\em 16th annual conference on neural information processing
  systems}, number CONF, pages 609--616, 2003.

\bibitem{snelson2006sparse}
E.~Snelson and Z.~Ghahramani.
\newblock Sparse gaussian processes using pseudo-inputs.
\newblock {\em Advances in neural information processing systems}, 18:1257,
  2006.

\bibitem{seeger2003bayesian}
Matthias Seeger.
\newblock Bayesian gaussian process models: Pac-bayesian generalisation error
  bounds and sparse approximations.
\newblock Technical report, University of Edinburgh, 2003.

\bibitem{sheth2015sparse}
Rishit Sheth, Yuyang Wang, and Roni Khardon.
\newblock Sparse variational inference for generalized gp models.
\newblock In {\em International Conference on Machine Learning}, pages
  1302--1311. PMLR, 2015.

\bibitem{GPforML}
Christopher~K Williams and Carl~Edward Rasmussen.
\newblock {\em Gaussian processes for machine learning}, volume~2.
\newblock MIT press Cambridge, MA, 2006.

\bibitem{matthews2017gpflow}
A.~Matthews et~al.
\newblock Gpflow: A gaussian process library using tensorflow.
\newblock {\em J. Mach. Learn. Res.}, 18(40):1--6, 2017.

\bibitem{abadi2016tensorflow}
M.~Abadi et~al.
\newblock Tensorflow: A system for large-scale machine learning.
\newblock In {\em 12th $\{$USENIX$\}$ symposium on operating systems design and
  implementation ($\{$OSDI$\}$ 16)}, pages 265--283, 2016.

\bibitem{quigley2009ros}
Morgan Quigley, Ken Conley, Brian Gerkey, Josh Faust, Tully Foote, Jeremy
  Leibs, Rob Wheeler, Andrew~Y Ng, et~al.
\newblock Ros: an open-source robot operating system.
\newblock In {\em ICRA workshop on open source software}, volume~3, page~5.
  Kobe, Japan, 2009.

\bibitem{hornung2013octomap}
A.~Hornung et~al.
\newblock Octomap: An efficient probabilistic 3d mapping framework based on
  octrees.
\newblock {\em Autonomous robots}, 34(3):189--206, 2013.

\end{thebibliography}









\end{document}